\documentclass[runningheads]{llncs}
\usepackage[T1]{fontenc}
\usepackage{graphicx,verbatim}
\usepackage{multirow}
\usepackage{amsfonts}
\usepackage{amsmath}
\usepackage{bbding}

\usepackage{booktabs}
\usepackage[table,xcdraw]{xcolor}

\begin{document}
\title{MedIQA: A Scalable Foundation Model for Prompt-Driven Medical Image Quality Assessment}
%

\author{Siyi Xun\inst{1} \and
Yue Sun\inst{1}\and
Jingkun Chen\inst{2} \and
Zitong Yu\inst{3} \and
Tong Tong\inst{4} \and
Xiaohong Liu\inst{5} \and
Mingxiang Wu\inst{6} \and
Tao Tan\inst{1}
}
\authorrunning{Xun et al.}

\institute{Faculty of Applied Sciences, Macao Polytechnic University, Macao, China \and
Department of Engineer Science, University of Oxford, Oxford, UK \and
Great Bay University, Dongguan, China \and
College of Physics and Information Engineering, Fuzhou University, Fuzhou, China \and
Shanghai Jiao Tong University, Shanghai, China \and
Department of Radiology, Shenzhen People’s Hospital, Shenzhen, China\\
Corresponding author: Tao Tan, \email{taotan@mpu.edu.mo}}

\maketitle              
\begin{abstract}

Rapid advances in medical imaging technology underscore the critical need for precise and automated image quality assessment (IQA) to ensure diagnostic accuracy. Existing medical IQA methods, however, struggle to generalize across diverse modalities and clinical scenarios. In response, we introduce MedIQA, the first comprehensive foundation model for medical IQA, designed to handle variability in image dimensions, modalities, anatomical regions, and types. We developed a large-scale multi-modality dataset with plentiful manually annotated quality scores to support this. Our model integrates a salient slice assessment module to focus on diagnostically relevant regions feature retrieval and employs an automatic prompt strategy that aligns upstream physical parameter pre-training with downstream expert annotation fine-tuning. Extensive experiments demonstrate that MedIQA significantly outperforms baselines in multiple downstream tasks, establishing a scalable framework for medical IQA and advancing diagnostic workflows and clinical decision-making.


\end{abstract}

\section{Introduction}

Medical image quality assessment (IQA) is critical for essential for reliable diagnosis. However, the heterogeneity of modalities, anatomical regions, and clinical scenarios poses significant challenges. Traditional IQA approaches, often based on handcrafted features or domain-specific models, struggle with generalization in various scenarios~\cite{1,2,3}. As medical imaging technologies become increasingly complex and data volumes surge, these limitations are further exacerbated.

Recent deep learning advances—especially foundation models pretrained on large-scale data—offer a powerful means to overcome these challenges. Foundation models, pretrained on large datasets, excel at learning universal representations that can be fine-tuned for specific tasks~\cite{4,5,6,31}. Their ability to generalize across domains and adapt to new tasks with minimal supervision makes them highly suitable for medical IQA. These models have demonstrated superior performance in tasks such as denoising, artifact detection, and quality scoring~\cite{7,8,9,29}.

Despite their promise, foundation models for medical IQA still face significant challenges: there is a scarcity of high-quality annotated datasets, a need for dynamic adaptation to varying imaging conditions, and difficulty integrating domain-specific knowledge into model architectures~\cite{10,11,12,28,30}. Moreover, the "black box" nature of these models limits their interpretability and clinical acceptance.

\begin{figure}

\includegraphics[width=\textwidth]{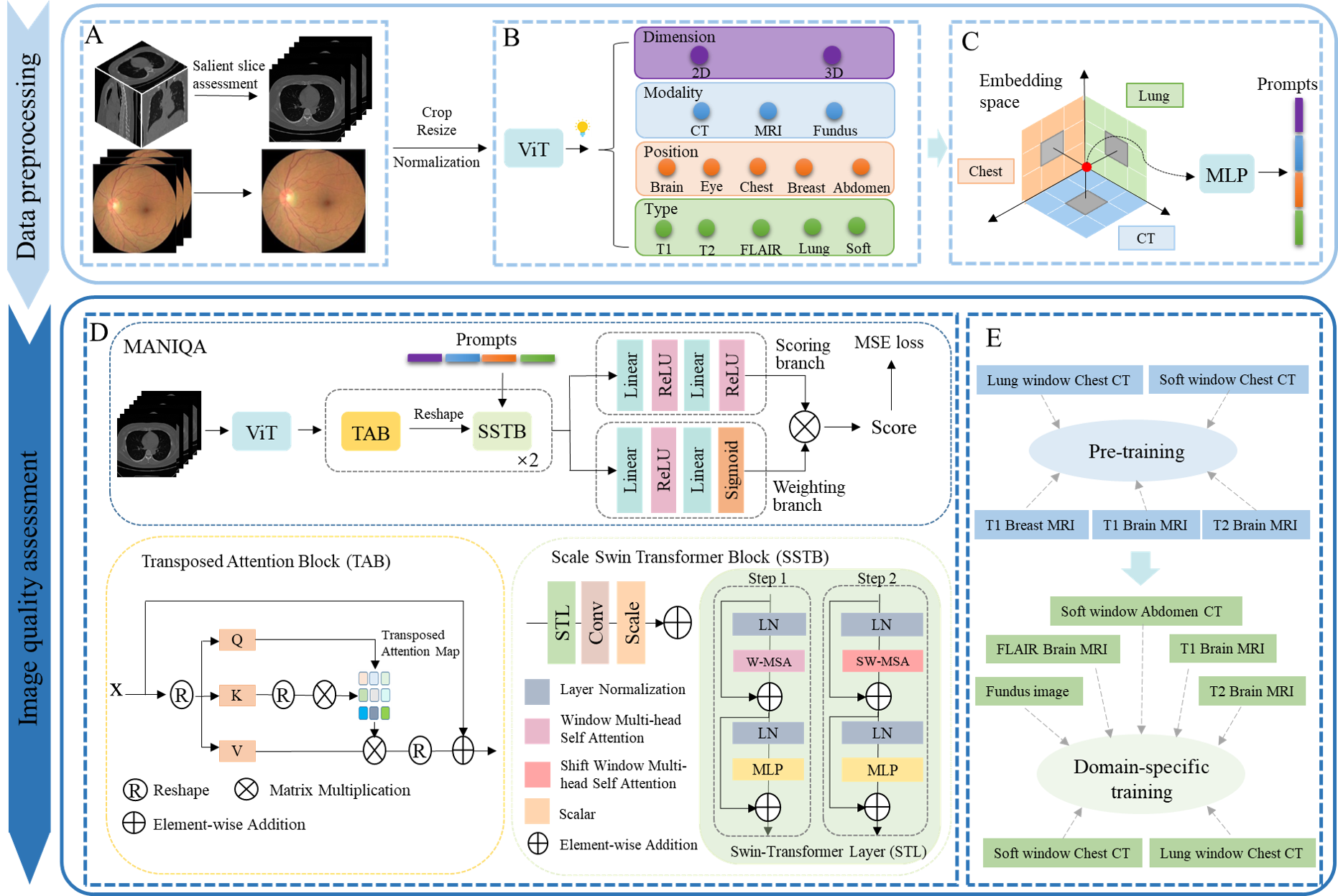}
\caption{Overview of the MedIQA workflow. (A) Salient slice assessment. (B) and (C) Prompts generation and encoding. (D) Backbone structure. (E) Training procedure.} \label{fig1}

\end{figure}

Based on these insights, we introduce MedIQA, a prompt-driven and scalable foundation model for medical IQA. Our contributions are as follows: \textit{(1)} We construct a large-scale MedIQA dataset of approximately 15k 2D and 3D radiographic scans, including CT, MRI, and other modalities, with high-quality expert annotations across various anatomical regions. \textit{(2)} We propose a salient slice assessment module to reduce redundant data and suppress background noise, allowing the model to focus on diagnostically relevant regions feature retrieval and enhancing both generalization and efficiency. \textit{(3)} We implement a two-stage training strategy: an upstream pre-training stage using physical parameters (e.g. dose, magnetic field strength) and a downstream fine-tuning stage with expert annotations. This approach creates an explicit link between objective physical characteristics and subjective quality assessment, thereby improving model interpretability. \textit{(4)} We integrate domain-specific imaging information (dimension, modality, position, and type) into an automated prompt strategy to ensure the model dynamically adapts to cross-modality multi-organ IQA tasks.

\section{Methodology}
\subsection{Dataset}

Existing medical IQA datasets are limited in scale, reliability, and diversity. To overcome these issues, we designed, proposed, and constructed a MedIQA dataset by integrating a large-scale annotated CT dataset and existing medical image datasets. The dataset is divided into a pretrain dataset (2,500 cases, including in-house chest CT and public MRI brain/breast datasets~\cite{14,15}) and a domain-specific dataset (12,545 cases, including expert-annotated Chest-CTIQA and public LDCTIQAC2023, ADNI MRI, and Kaggle DR datasets~\cite{17,18,19}). The pre-training dataset and Chest-CTIQA are 3D data, while LDCTIQAC2023 and Kaggle DR are 2D data. Chest-CTIQA is the first large chest CT IQA dataset (10 readers per volume). For ADNI MRI, in order to ensure the quantity and balance of data, we selected different 2D images from the 3D volume to construct the MRI dataset. The pretrain data labels were generated by extracting dose (mAs) and magnetic field strength (Tesla) from image physical parameters, with label values positively correlated with image quality~\cite{13,16}. All domain-specific images were annotated by radiologists or trained professionals. All datasets were preprocessed for consistency, and normalized for training. MedIQA is the first multimodal IQA dataset addressing diverse quality assessment needs. Fig.2 shows dataset examples, quantities, and distributions.

\begin{figure}

\includegraphics[width=\textwidth]{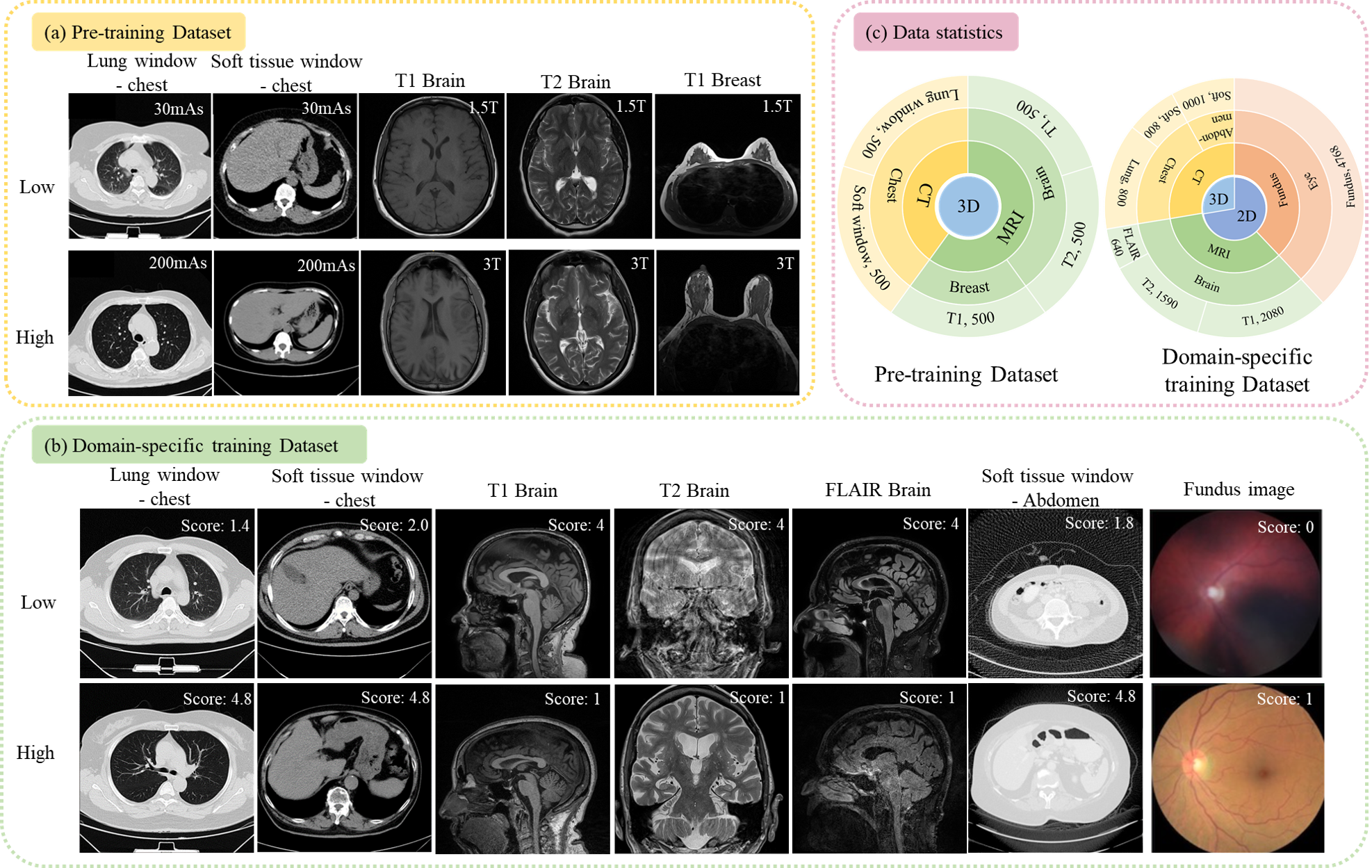}
\caption{Examples, quantities, and distributions of the MedIQA-dataset.} \label{fig2}

\end{figure}

\subsection{Model}

The overall architecture of the model is designed to achieve comprehensive medical IQA, as illustrated in Fig.1. First, the image input module receives and preprocesses input images. For 3D volume, the salient slice assessment module extracts seven salient slices from the sequence, focusing on diagnostic regions feature retrieval. Next, the pre-trained Vision Transformer (ViT) classifier generates encoded prompts for dynamic adaptation to different imaging conditions while matching upstream and downstream tasks. The main framework of the model uses MANIQA~\cite{20} as the backbone network for feature extraction and quality assessment. Finally, the model outputs the overall image quality score.

\textbf{Salient slice assessment for feature retrieval.} Due to the minimal quality differences between adjacent slices, continuous sampling often results in redundant data. Therefore, in order to focus on the region of interest while reducing computational complexity, we select seven salient slices from each volume for feature retrieval (Fig.1.A). Specifically, the 3D medical volume $V \in \mathbb{R} ^{H\times W\times D} $ is divided into seven regions $R_{i}  = S[v_{i-1}:v_{i}  ], i = 1,2,...,7$ along the $Z$ axis by removing the irrelevant slice that does not contain any region of interest. The middle slice $s_{i} = R_{i} \left [  \left \lfloor \frac{\left | R_{i}  \right | }{2}  \right \rfloor  \right ] $ is selected from each region to ensure uniform sampling constraints globally while covering the local critical slice that contains the diagnostic region. All images were min-max normalized to align the intensity distribution, and image sizes were normalized to 224*224 for consistency.

\textbf{Upstream physical parameters-driven foundation model learning.} Physical parameters $p\in \mathbb{R}^{k}$ such as dose or magnetic field strength are pretrained to help the model learn the effects of these parameters on underlying image features $f\in \mathbb{R}^{m}$ (noise, contrast, resolution). The objective function is defined by $\underset{\theta }{min} \mathcal{L} _{pre } = \mathbb{E} _{(\tau, p )} \left [ \left \| g_{\varphi} (E_{\theta } (\tau)) -p\right \| _{2}^{2}  \right ] $, where the encoder $E_{\theta} : \tau \to f$, the parameter is $\theta$, $g_{\varphi} : f \to p$ is the parameter prediction head, and the explicit association between the feature $f$ and $p$ is constrained by the mean square error (MSE). The explicit association formed by the "parameter → feature" mapping established in the pre-training stage provides a physical basis for the subsequent quality assessment. In the fine-tuning stage, the intermediate features generated by pre-training and strongly related to physical parameters will be reused to make the model decision-making process more transparent.

\textbf{Prompt-based upstream and downstream matching.} Prompts include: dimensional ($p_{dim}$), modality ($p_{mod}$), region ($p_{reg}$), and type ($p_{type}$). Dimensional prompts are derived from input dimensions, while others are generated using a pre-trained ViT for classification. A 12-layer, 12-head ViT with a size of 3072 is employed as the decoder (Fig.1.B). For integration, four one-hot encoded prompts are concatenated and projected into Swin Transformer Layers (STL) via a fully connected (FC) layer (Fig.1.C). Prompts are added to the feature vector at each decoder layer, enabling dynamic adaptation. Each layer has an independent FC layer, allowing task updates without new branches. The implementation is, $y =  x + FC_{i} (concat(p_{dim}, p_{mod}, p_{reg}, p_{type} )), i = 1,2$. Prompts contribute the same amount in each STL layer. Prompt strategy matches upstream physical parameters-driven foundation model learning with downstream expert annotation-driven domain-specific knowledge learning to achieve dual supervision of expert annotation and physical characteristics.

\textbf{Quality assessment.} Given the lack of high-quality reference images in medical IQA, no-reference IQA (NR-IQA) has become the optimal choice. Therefore, we adopt MANIQA, a state-of-the-art NR-IQA model, as our backbone (Fig.1.D). MANIQA uses ViT extracts features $F \in \mathbb{R}^{b\times \sum_{i}^{} C_{i} \times H_{i}W_{i}   } $ sent to Transposed Attention Block (TAB) and Scale Swin Transformer Block (SSTB) to implement multi-dimensional attention mechanisms in both channel and spatial. The final score is given by the dual-branch prediction module of the scoring ($s$) branch and the weighting ($w$) branch. For 2D images, predict quality score is computed as $q = \frac{\sum_{0<i<N}^{} w_{i} \times s_{i} }{\sum_{0<i<N}^{} w_{i}} $, where $N$ denotes the number of patches for one image. For 3D volumes, features are extracted from seven salient slices, slice-level scores $\mathbf{q}  = [q_{1},  q_{2}, ...,q_{7}]$ are obtained by dual-branch structure, corresponding weights $\mathbf{\bar{w} }  = [\bar{w}_{1},  \bar{w}_{2}, ...,\bar{w}_{7}]$ are generated by linear layer, and the final image quality score $Q$ is calculated as $Q =  \sum_{i = 1}^{7} \bar{w} _{i} q_{i} $ . These weights dynamically adjust based on the importance of each slice, enhancing the model's precision and sensitivity to local quality variations.

\subsection{Model Training Procedure}

\textbf{Training process: }During the pre-training stage, we use the pret-raining dataset and labels derived from DICOM tags related to image quality. In domain-specific training stage, we use domain-specific datasets (Fig.1.E). The data were manually annotated by experienced radiologists and trained professionals. Both stages share consistent training settings, differing only in the training data and labels.

\textbf{Loss Function}: The MSE loss function is used to measure the difference between predicted quality scores and true annotations. MSE ensures minimized prediction errors and alignment with expert annotations, while its smoothness promotes stable convergence during optimization. The MSE loss is calculated as $L_{MSE} = \frac{1}{n}\sum (x_{i}-y_{i} )^{2}$, where $n$ is the number of samples, $x_i$ and $y_i$ is the true value and the predicted value of the model.

\section{Experiments}
\subsection{Implementation Details and Metrics}

Our experiments are implemented on an Intel(R) Xeon(R) W2245 CPU @ 3.90GHz and an NVIDIA RTX A6000 GPU with Python 3.9 and PyTorch 1.10 for training and testing. Hyperparameters included a learning rate of 1e-5, a batch size of 1, and 50 training epochs, optimized using the Adam optimizer. The dataset was split into training, validation, and test sets in an 8:1:1 ratio across diverse data distributions.

\begin{table}[]
\caption{Analyse the performance on the upstream and downstream tasks. Our-s and Our-e indicate the single baseline and the ensemble model. Best in red and second in blue. $ \mathcal{S}$, SRCC; $\mathcal{P}$, PLCC; $\mathcal{R}$, RMSE. }
\resizebox{\textwidth}{!}{
\begin{tabular}{@{}lcccccccccccccccccc@{}}
\toprule
                                               & \multicolumn{18}{c}{\cellcolor[HTML]{ECF4FF}Upstream -- pretraining dataset}                                                                                                                                                                                                                                                                                                                                                                                                                                                                                                                   \\ \midrule
\multicolumn{1}{c}{}                           & \multicolumn{3}{c}{Test}                                                                      & \multicolumn{3}{c}{Lung window}                                                                & \multicolumn{3}{c}{Soft window}                                                               & \multicolumn{3}{c}{Brain T1}                                                                  & \multicolumn{3}{c}{Brain T2}                                                                  & \multicolumn{3}{c}{Breast T1}                                                                 \\
\multicolumn{1}{c}{\multirow{-2}{*}{Backbone}} & $\mathcal{S}$ ↑                        & $\mathcal{P}$ ↑                        & $\mathcal{R}$ ↓                        & $\mathcal{S}$ ↑                        & $\mathcal{P}$ ↑                        & $\mathcal{R}$ ↓                         & $\mathcal{S}$ ↑                        & $\mathcal{P}$ ↑                        & $\mathcal{R}$ ↓                        & $\mathcal{S}$ ↑                        & $\mathcal{P}$ ↑                        & $\mathcal{R}$ ↓                        & $\mathcal{S}$ ↑                        & $\mathcal{P}$ ↑                        & $\mathcal{R}$ ↓                        & $\mathcal{S}$ ↑                        & $\mathcal{P}$ ↑                        & $\mathcal{R}$ ↓                        \\
VGG                                            & 0.6623                        & 0.6744                        & {\color[HTML]{3531FF} 0.2838} & 0.6857                        & 0.6972                        & 0.2141                         & 0.7054                        & {\color[HTML]{3531FF} 0.7247} & {\color[HTML]{3531FF} 0.2149} & {\color[HTML]{3531FF} 0.6838} & 0.6770                        & {\color[HTML]{3531FF} 0.2679} & {\color[HTML]{3531FF} 0.5914} & {\color[HTML]{3531FF} 0.6142} & 0.4153                        & 0.6452                        & 0.6589                        & {\color[HTML]{3531FF} 0.3062} \\
Resnet                                         & 0.5413                        & 0.5506                        & 0.2994                        & 0.6283                        & 0.6438                        & 0.2112                         & 0.2988                        & 0.2848                        & 0.2569                        & 0.5945                        & 0.6129                        & 0.3112                        & 0.5559                        & 0.5901                        & {\color[HTML]{3531FF} 0.4083} & 0.6290                        & 0.6187                        & 0.3121                        \\
Swin-transformer                               & {\color[HTML]{3531FF} 0.6626} & {\color[HTML]{3531FF} 0.6758} & 0.2842                        & 0.7034                        & 0.7221                        & {\color[HTML]{3531FF} 0.2102}  & 0.7018                        & 0.7021                        & 0.2175                        & 0.6619                        & {\color[HTML]{3531FF} 0.6771} & 0.2771                        & 0.5566                        & 0.5774                        & 0.4137                        & {\color[HTML]{3531FF} 0.6886} & {\color[HTML]{3531FF} 0.6792} & 0.3086                        \\
ViT                                            & 0.5852                        & 0.5858                        & 0.3044                        & 0.7118                        & 0.7021                        & 0.2175                         & 0.6591                        & 0.6498                        & 0.2339                        & 0.5396                        & 0.5515                        & 0.3313                        & 0.4193                        & 0.4378                        & 0.4284                        & 0.5993                        & 0.5887                        & 0.3211                        \\
DeepViT                                        & 0.5402                        & 0.5432                        & 0.3108                        & 0.5490                        & 0.5644                        & 0.2396                         & 0.6203                        & 0.6054                        & 0.2376                        & 0.5213                        & 0.5339                        & 0.3338                        & 0.4127                        & 0.4422                        & 0.4327                        & 0.5968                        & 0.5857                        & 0.3217                        \\
CNNIQAnet                                      & 0.5486                        & 0.5266                        & 0.3202                        & 0.5549                        & 0.5466                        & 0.2471                         & 0.6096                        & 0.5876                        & 0.2404                        & 0.5382                        & 0.5032                        & 0.3423                        & 0.4957                        & 0.5098                        & 0.4499                        & 0.5434                        & 0.4851                        & 0.3404                        \\
WaDIQaM                                        & 0.6478                        & 0.6298                        & 0.2998                        & {\color[HTML]{3531FF} 0.7152} & {\color[HTML]{3531FF} 0.7225} & 0.2103 & {\color[HTML]{3531FF} 0.7091} & 0.6891                        & 0.2196                        & 0.6154                        & 0.5930                        & 0.3192                        & 0.5873                        & 0.5546                        & 0.4135                        & 0.6094                        & 0.5657                        & 0.3173                        \\
Ours                                           & {\color[HTML]{FE0000} 0.7770} & {\color[HTML]{FE0000} 0.8110} & {\color[HTML]{FE0000} 0.2609} & {\color[HTML]{FE0000} 0.7316} & {\color[HTML]{FE0000} 0.7363} & {\color[HTML]{FE0000} 0.2029}  & {\color[HTML]{FE0000} 0.7114} & {\color[HTML]{FE0000} 0.7674} & {\color[HTML]{FE0000} 0.2031} & {\color[HTML]{FE0000} 0.7891} & {\color[HTML]{FE0000} 0.8739} & {\color[HTML]{FE0000} 0.2443} & {\color[HTML]{FE0000} 0.6828} & {\color[HTML]{FE0000} 0.6689} & {\color[HTML]{FE0000} 0.3843} & {\color[HTML]{FE0000} 0.8148} & {\color[HTML]{FE0000} 0.9246} & {\color[HTML]{FE0000} 0.1957} \\
                                               & \multicolumn{18}{c}{\cellcolor[HTML]{EFFDEB}Downstream -- domain-specific datasets}                                                                                                                                                                                                                                                                                                                                                                                                                                                                                                            \\
\multicolumn{1}{c}{}                           & \multicolumn{3}{c}{Chest-CTIQA}                                                               & \multicolumn{3}{c}{Brain-T1}                                                                   & \multicolumn{3}{c}{Brain-T2}                                                                  & \multicolumn{3}{c}{Brain-FLAIR}                                                               & \multicolumn{3}{c}{Fundus}                                                                    & \multicolumn{3}{c}{LDCTIQAC2023}                                                              \\
\multicolumn{1}{c}{\multirow{-2}{*}{Backbone}} & $\mathcal{S}$ ↑                        & $\mathcal{P}$ ↑                        & $\mathcal{R}$ ↓                        & $\mathcal{S}$ ↑                        & $\mathcal{P}$ ↑                        & $\mathcal{R}$ ↓                         & $\mathcal{S}$ ↑                        & $\mathcal{P}$ ↑                        & $\mathcal{R}$ ↓                        & $\mathcal{S}$ ↑                        & $\mathcal{P}$ ↑                        & $\mathcal{R}$ ↓                        & $\mathcal{S}$ ↑                        & $\mathcal{P}$ ↑                        & $\mathcal{R}$ ↓                        & $\mathcal{S}$ ↑                        & $\mathcal{P}$ ↑                        & $\mathcal{R}$ ↓                        \\
VGG                                            & 0.4166                        & 0.3686                        & 0.2670                        & 0.6172                        & 0.6127                        & 0.2103                         & 0.5313                        & 0.5394                        & 0.3419                        & 0.6283                        & 0.6214                        & 0.3345                        & 0.7368                        & 0.6923                        & 0.2899                        & 0.8350                        & 0.8298                        & 0.1609                        \\
Resnet                                         & 0.3902                        & 0.3976                        & 0.2026                        & 0.5719                        & 0.5284                        & 0.2571                         & 0.4919                        & 0.5039                        & 0.3467                        & 0.5202                        & 0.5214                        & 0.3556                        & 0.7119                        & 0.6897                        & 0.3772                        & 0.7972                        & 0.7775                        & 0.1599                        \\
Swin-transformer                               & 0.4741                        & 0.5004                        & 0.2232                        & 0.8026                        & 0.8694                        & 0.2137                         & 0.6968                        & 0.7012                        & 0.3749                        & 0.6968                        & 0.7012                        & 0.2849                        & 0.8118                        & 0.7755                        & 0.3068                        & 0.9013                        & 0.9009                        & 0.0922                        \\
ViT                                            & 0.4044                        & 0.4225                        & 0.2040                        & 0.6373                        & 0.6797                        & 0.2589                         & 0.6015                        & 0.5631                        & 0.3051                        & 0.5659                        & 0.5664                        & 0.3364                        & 0.7156                        & 0.6717                        & 0.3777                        & 0.8127                        & 0.7432                        & 0.1124                        \\
DeepViT                                        & 0.4592                        & 0.4697                        & 0.2964                        & 0.7988                        & 0.8646                        & 0.1822                         & 0.6585                        & 0.6582                        & 0.2775                        & 0.6131                        & 0.5879                        & 0.3203                        & 0.7368                        & 0.6923                        & 0.3899                        & 0.8342                        & 0.8336                        & 0.1257                        \\
CNNIQAnet                                      & 0.4641                        & 0.4990                        & 0.2530                        & 0.6547                        & 0.6567                        & 0.2161                         & 0.6517                        & 0.6369                        & 0.3027                        & 0.6323                        & 0.6286                        & 0.3429                        & 0.7842                        & 0.7734                        & 0.3002                        & 0.8661                        & 0.8866                        & 0.1036                        \\
WaDIQaM                                        & 0.4835                        & 0.4810                        & 0.1945                        & 0.7119                        & 0.6897                        & 0.2172                         & 0.7056                        & 0.6717                        & 0.2777                        & 0.6585                        & 0.6582                        & 0.3275                        & 0.7672                        & 0.7775                        & 0.2519                        & 0.8798                        & 0.8583                        & 0.1213                        \\
Ours-s                                         & {\color[HTML]{3531FF} 0.4875} & {\color[HTML]{3531FF} 0.5255} & {\color[HTML]{3531FF} 0.1661} & {\color[HTML]{3531FF} 0.8659} & {\color[HTML]{3531FF} 0.8905} & {\color[HTML]{3531FF} 0.1507}  & {\color[HTML]{3531FF} 0.7058} & {\color[HTML]{3531FF} 0.7035} & {\color[HTML]{3531FF} 0.2731} & {\color[HTML]{3531FF} 0.7177} & {\color[HTML]{3531FF} 0.7184} & {\color[HTML]{3531FF} 0.2798} & {\color[HTML]{3531FF} 0.8329} & {\color[HTML]{3531FF} 0.9144} & {\color[HTML]{3531FF} 0.2054} & {\color[HTML]{3531FF} 0.9761} & {\color[HTML]{3531FF} 0.9759} & {\color[HTML]{3531FF} 0.0631} \\
Ours-e                                         & {\color[HTML]{FE0000} 0.7070} & {\color[HTML]{FE0000} 0.7455} & {\color[HTML]{FE0000} 0.1276} & {\color[HTML]{FE0000} 0.8681} & {\color[HTML]{FE0000} 0.8985} & {\color[HTML]{FE0000} 0.1196}  & {\color[HTML]{FE0000} 0.8861} & {\color[HTML]{FE0000} 0.8912} & {\color[HTML]{FE0000} 0.1696} & {\color[HTML]{FE0000} 0.7654} & {\color[HTML]{FE0000} 0.7578} & {\color[HTML]{FE0000} 0.2657} & {\color[HTML]{FE0000} 0.8504} & {\color[HTML]{FE0000} 0.9320} & {\color[HTML]{FE0000} 0.1837} & {\color[HTML]{FE0000} 0.9764} & {\color[HTML]{FE0000} 0.9762} & {\color[HTML]{FE0000} 0.0618} \\ \bottomrule
\end{tabular}
}
\end{table}

The performance of IQA tasks was evaluated by Spearman rank-order correlation coefficient (SRCC), Pearson linear correlation coefficient (PLCC) and Root mean square error (RMSE). SRCC and PLCC measure the monotonicity and linear correlation of the model, while RMSE assesses the consistency of the model's predictions.

\subsection{Experimental Results and Analysis}

\textbf{Classification experiments. }To enable automatic prompt generation, we trained VGG and ViT models using additional classification data and evaluated them on the MedIQA dataset. Experimental results show that VGG achieved an average test accuracy of 0.9445, while ViT achieved 0.9969. Meanwhile, ViT's self-attention-based design allowed for higher performance with fewer parameters (86M \textit{VS} 138M). Therefore, we selected ViT for prompt generation.

\textbf{Upstream foundation experiments.} The proposed model was trained and tested on pre-training dataset as well as five sub-benchmarks (chest lung window benchmark, chest soft tissue window benchmark, brain T1 benchmark, brain T2 benchmark and breast T1 benchmark). Table 1 shows the results of our model and other methods. The experimental results show that the proposed model can learn quality features of different modality images, accurately predict quality-related parameters on different benchmarks, and provide a good basis for training downstream tasks.

\textbf{Downstream tasks.} In domain-specific task experiments, we evaluated the model's performance on six benchmarks: 3D chest CT benchmark, 2D brain T1, T2, and FLAIR MRI benchmarks, 2D fundus image benchmark, and 2D synthetic abdominal CT benchmark. Results (Table 1) show that the model performed relatively poorly on 3D chest CT data due to the complexity of high-dimensional data. In contrast, synthetic abdominal CT data achieved the best results, as the significant quality variations enabled the model to learn assessment features more effectively. T1 and T2 MRI data showed stable performance, while FLAIR MRI data underperformed due to a lack of pre-training data information. Fundus image assessment results were moderate, likely due to resolution and lighting limitations. Future work should focus on optimizing 3D data handling and leveraging synthetic data for pre-training.

\textbf{Performance Comparisons.} To validate the effectiveness of MedIQA, we evaluated it against other models, including VGG~\cite{22}, ResNet~\cite{23}, ViT, Swin-Transformer~\cite{24}, DeepViT~\cite{25}, CNNIQAnet~\cite{26}, and WaDIQaM~\cite{27}. Results (Table 1) demonstrate that our model outperformed all others in image quality assessment tasks. In upstream tasks, the average result of our model (0.7511, 0.7970, 0.2485) is significantly improved (+0.2027, +0.2706, -0.0748) compared with the average result of CNNIQAnet (0.5484, 0.5264, 0.3233). For downstream tasks, the average results of our model (0.8422, 0.8668, 0.1546) improved by 7.79\% and 7.88\% over the average results of baseline (0.7643, 0.7880, 0.1897). It is also significantly higher than the average result of ResNet (+0.2617, +0.2971, -0.1285). The experimental results show that compared with classical models (VGG and ResNet), our model's multi-dimensional attention mechanisms and salient slice assessment modules better retrieve local and global features, overcoming traditional CNN limitations and significantly improving accuracy. Additionally, our model outperformed other transformer-based models by enhancing generalization through pretraining and prompt strategies, while capturing finer details via salient slice assessment module. It also surpassed specialized IQA models like CNNIQAnet and WaDIQaM. These results validate the effectiveness of our novel designs, offering an efficient solution for IQA tasks.

\begin{table}[]
\caption{Ablation study on downstream tasks. Only 3D images use the salient slice assessment module. Best in red and second in blue.}

\resizebox{\textwidth}{!}{
\begin{tabular}{@{}cccccccccccccccccccccc@{}}
\toprule
                        & \multicolumn{3}{c}{Module} & \multicolumn{3}{c}{Chest-CTIQA(3D)}                                                               & \multicolumn{3}{c}{Brain-T1(2D)}                                                                  & \multicolumn{3}{c}{Brain-T2(2D)}                                                                  & \multicolumn{3}{c}{Brain-FLAIR(2D)}                                                               & \multicolumn{3}{c}{Fundus(2D)}                                                                    & \multicolumn{3}{c}{LDCTIQAC2023(2D)}                                                              \\ \cmidrule(l){2-22} 
\multirow{-2}{*}{Model} & PT      & PM      & SS     & $\mathcal{S}$  ↑                        & $\mathcal{P}$  ↑                        & $\mathcal{R}$  ↓                        & $\mathcal{S}$  ↑                        & $\mathcal{P}$  ↑                        & $\mathcal{R}$  ↓                        & $\mathcal{S}$  ↑                        & $\mathcal{P}$  ↑                        & $\mathcal{R}$  ↓                        & $\mathcal{S}$  ↑                        & $\mathcal{P}$  ↑                        & $\mathcal{R}$  ↓                        & $\mathcal{S}$  ↑                        & $\mathcal{P}$  ↑                        & $\mathcal{R}$  ↓                        & $\mathcal{S}$  ↑                        & $\mathcal{P}$  ↑                        & $\mathcal{R}$  ↓                        \\ \cmidrule(r){1-22}
1                       & \XSolidBrush       & \XSolidBrush       & \XSolidBrush      & 0.4875                        & 0.5255                        & 0.1661                        & 0.8659                        & 0.8905                        & 0.1507                        & 0.7058                        & 0.7035                        & 0.2731                        & 0.7177                        & 0.7184                        & 0.2798                        & 0.8329                        & 0.9144                        & 0.2054                        & {\color[HTML]{3531FF} 0.9761}                        & {\color[HTML]{3531FF} 0.9759}                          & 0.0631                        \\
2                       & \Checkmark      & \XSolidBrush       & \XSolidBrush      & 0.5347                        & 0.5541                        & 0.1608                        & {\color[HTML]{FE0000} 0.8706} & {\color[HTML]{FE0000} 0.8978} & {\color[HTML]{3531FF} 0.1261}                          & {\color[HTML]{3531FF} 0.8549}                         & {\color[HTML]{3531FF} 0.8559}                         & {\color[HTML]{3531FF} 0.1936 }                        & {\color[HTML]{FE0000} 0.8294} & {\color[HTML]{FE0000} 0.8247} & {\color[HTML]{FE0000} 0.2203} & {\color[HTML]{3531FF} 0.8345}                        & {\color[HTML]{FE0000} 0.9438} & {\color[HTML]{FE0000} 0.1668} & 0.9742                        & 0.9757                        & {\color[HTML]{FE0000} 0.0589} \\
3                       & \Checkmark      & \Checkmark      & \XSolidBrush      & {\color[HTML]{3531FF} 0.5464}                       & {\color[HTML]{3531FF} 0.5927}                           &     {\color[HTML]{3531FF} 0.1561}                     & {\color[HTML]{3531FF} 0.8681}                         & {\color[HTML]{3531FF} 0.8985}                        & {\color[HTML]{FE0000} 0.1196} & {\color[HTML]{FE0000} 0.8861} & {\color[HTML]{FE0000} 0.8912} & {\color[HTML]{FE0000} 0.1696} & {\color[HTML]{3531FF} 0.7654}                         & {\color[HTML]{3531FF} 0.7578}                         & {\color[HTML]{3531FF} 0.2657}                         & {\color[HTML]{FE0000} 0.8504} & {\color[HTML]{3531FF} 0.9320}                         & {\color[HTML]{3531FF} 0.1837}                        & {\color[HTML]{FE0000} 0.9764} & {\color[HTML]{FE0000} 0.9762} & {\color[HTML]{3531FF} 0.0618}                         \\
4                       & \Checkmark      & \Checkmark      & \Checkmark     & {\color[HTML]{FE0000} 0.7070} & {\color[HTML]{FE0000} 0.7455} & {\color[HTML]{FE0000} 0.1276} & /     & /     & /     & /     & /     & /     & /     & /     & /     & /     & /     & /     & /     & /     & /     \\ \bottomrule
\end{tabular}
}
\end{table}

\textbf{Ablation Study.} We evaluated the impact of pretraining (PT), prompt strategies (PM), and salient slice assessment (SS) performance on domain-specific downstream tasks. Results (Table 2) show that the modules' effects vary by data type. For 3D chest CT data, all modules improved performance, with the salient slice assessment module significantly enhancing 3D feature learning (+0.2195, +0.2200, -0.0385). For 2D T2 data, all modules had positive effects (+0.1803, +0.1877, -0.1035), as the prompt strategy and pretraining helped capture quality features. However, for 2D FLAIR data, the prompt strategy caused performance degradation, likely due to mismatches between FLAIR's unique quality characteristics and the prompt strategy's design, leading the model to learn irrelevant features. For 2D T1, 2D fundus, and 2D synthetic CT data, additional modules had minimal impact, as these datasets' simplicity or consistency already enabled strong performance. Future work should optimize prompt designs for different modalities to improve generalization and performance.

\section{Discussion}

By integrating large-scale cross-modality MedIQA dataset, prompt strategy and salient slice assessment module and upstream and downstream matching, MedIQA captures both global and local quality features, ensuring robust assessments.  Compared to other methods, our model shows significant improvements and provides a scalable framework for medical IQA. In addition, preliminary experiments revealed that CT image quality affects AI detection of lung nodules. Thus, the relationship between medical image quality and disease detection rates will be a focus of our future research.

Despite promising results, the study has limitations. First, pretraining data may not fully capture variability across modalities or clinical scenarios, necessitating more annotated data or unsupervised learning methods. Second, the prompt strategy, while effective, relies heavily on high-quality prompt design, requiring further optimization for diverse tasks. Third, salient slice assessment module may miss subtle quality changes in long sequences, potentially underperforming in extreme conditions (e.g., excessive noise or missing images). Future research will focus on: (1) Expanding datasets to include diverse modalities and scenarios for better generalizability; (2) Developing interpretable architectures to build clinician trust; and (3) Integrating the model into clinical workflows and validating its impact on diagnostic accuracy and efficiency.

\section{Conclusion}

In this paper, we proposed a scalable foundation model for medical IQA. First, we constructed the MedIQA dataset, a large-scale, multi-modal, and multi-organ dataset with DICOM tags and plentiful manually annotated quality scores, providing a robust foundation for model learning. Second, we designed a salient slice assessment module to focus on diagnostically relevant regions and enhance efficiency, and implement a two-stage training strategy to bridge physical parameters with expert annotations, improving interpretability. We also designed domain-specific automated prompts for cross-modality multi-organ IQA tasks. Experimental results demonstrate superior performance across multiple IQA benchmarks and solved the limitations of traditional methods. Our approach provides a scalable solution for clinical applications. Future research will focus on mitigating data scarcity, optimizing prompt strategies, and refining salient slice assessment to further enhance the model's practicality and applicability.

\bibliographystyle{splncs04_unsort}
\bibliography{main}

\end{document}